\documentclass[times, review, 10pt]{elsarticle}

\usepackage{setspace}
\usepackage[numbers]{natbib}
\usepackage{amsmath,amsfonts}
\usepackage{color}
\usepackage{amssymb}
\usepackage{amsmath}

\usepackage{booktabs}
\usepackage{multirow}
\usepackage{amssymb}
\usepackage{array}
\usepackage{paralist}       
\usepackage{amsmath} 
\usepackage{url}
\usepackage{stfloats}
\usepackage{subcaption}

\usepackage{amsmath,amsfonts}
\usepackage{algorithm}
\usepackage{algorithmic}
\usepackage{array}
\usepackage{textcomp}
\usepackage{stfloats}
\usepackage{url}
\usepackage{verbatim}
\usepackage[table]{xcolor}
\usepackage{graphicx}
\usepackage{booktabs}
\usepackage{multirow}
\usepackage{bbding}
\usepackage{caption}
\usepackage{subcaption}
\usepackage{tabularx, booktabs}

\newcommand{\etal}{\textit{et}~\textit{al}.} 
\journal{Elsevier}

\begin{document}
\definecolor{gray}{rgb}{0.5, 0.5, 0.5}  
\begin{frontmatter}



\title{Exploring the Temporal Consistency for
Point-Level Weakly-Supervised Temporal Action Localization}






\author[1]{Yunchuan Ma}
\ead{mayunchuan23@mails.ucas.ac.cn}
\author[1]{Laiyun Qing\corref{cor}}
\author[1]{Guorong Li}
\author[1]{Yuqing Liu}
\author[2]{Yuankai Qi}
\author[1]{Qingming Huang}
\cortext[cor]{Corresponding author}

\address[1]{University of Chinese Academy of Sciences, Beijing, 100190, China}
\address[2]{Macquarie University}

\begin{abstract}
Point-supervised Temporal Action Localization (PTAL) adopts a lightly frame-annotated paradigm (\textit{i.e.}, labeling only a single frame per action instance) to train a model to effectively locate action instances within untrimmed videos.
Most existing  approaches 
design the task head of models with only a
point-supervised snippet-level classification, without explicit modeling of understanding temporal relationships among frames of an action.
However, understanding the temporal relationships of frames is crucial because it can help a model understand how an action is defined and therefore benefits the localization of complete action instances.
To this end, in this paper, we design a multi-task learning framework that fully utilizes point supervision to boost the model's temporal understanding capability for action localization.
Specifically, we design three self-supervised temporal understanding tasks:
(i) Action Completion, (ii) Action Order Understanding, and (iii) Action Regularity Understanding. 
These tasks help a model understand the temporal consistency of actions across videos.
To the best of our knowledge, this is the first attempt to explicitly explore
temporal consistency
for point-supervision action localization.
Extensive experimental results on four benchmark datasets 
demonstrate the effectiveness of the proposed method compared to several state-of-the-art approaches.

\end{abstract}

\end{frontmatter}



\begin{figure}[!t]
\centering
\includegraphics[width=\linewidth]{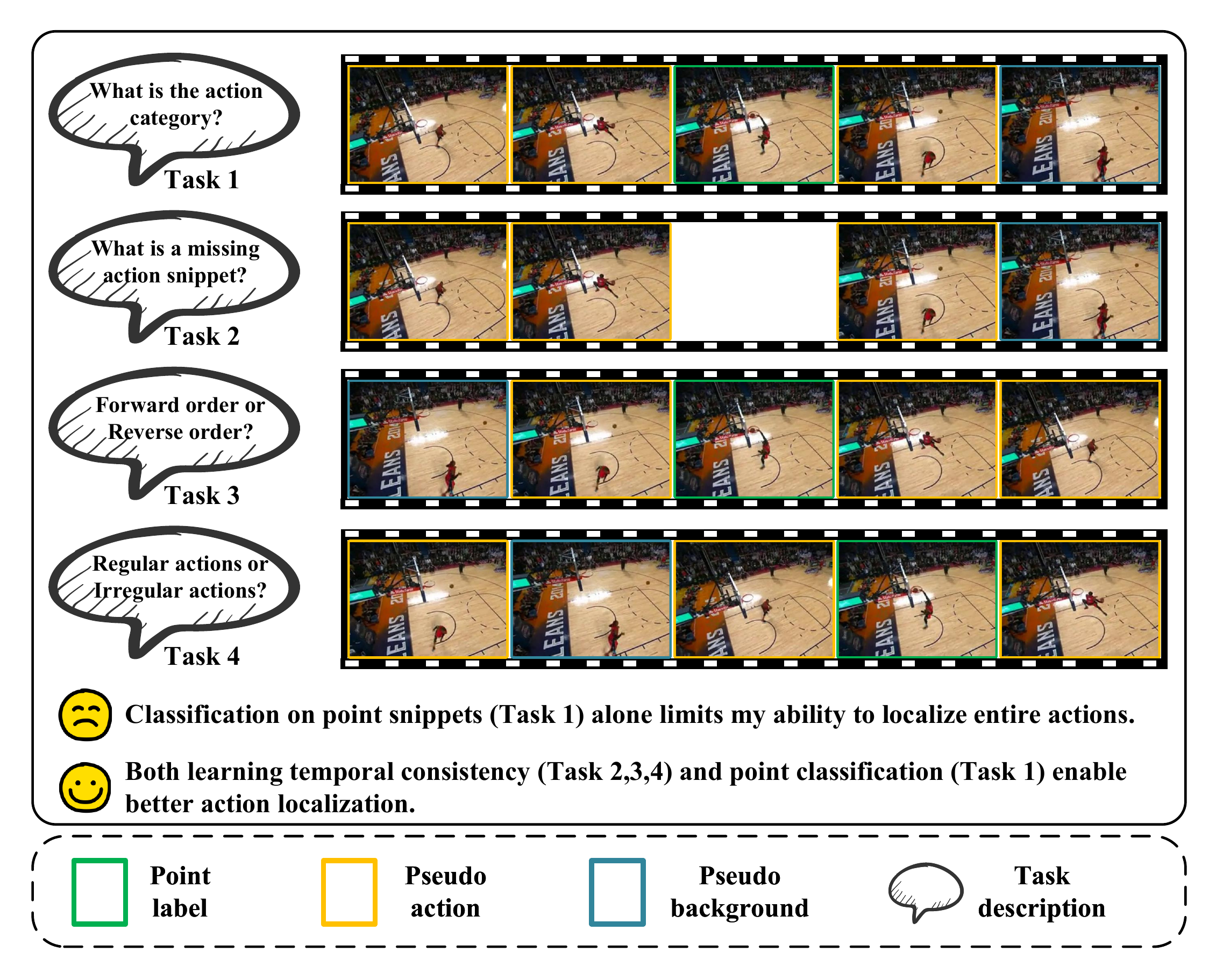}
\caption{Previous methods primarily use point annotations to supervise the action classification of labeled snippets (Task 1). 
We improve upon this by incorporating three point-based self-supervised tasks (Tasks 2, 3, and 4) that enable the model to better capture temporal consistency and understand action sequences, thereby boosting action localization.
}
\label{fig:insight}

\end{figure}

\section{Introduction}
Temporal action localization (TAL) plays a critical role in the video understanding field~\cite{pr1}, aiming to predict the start and end timestamps of each action instance in the given untrimmed videos.
Thanks to accurate action boundary labeling, conventional fully-supervised temporal action localization (FTAL)~\cite{pr4,pr5} methods have achieved remarkable progress. Nevertheless, the frame-level annotation of action instances is laborious and time-consuming.
To reduce the cost of annotating, researchers have increasingly turned to weakly-supervised temporal action localization (WTAL)~\cite{pr2,pr3,NoCo}, which uses only video-level labels. Although significant progress has been made, its performance remains limited due to the lack of fine-grained annotations.

To better balance annotation cost and model performance, point-supervised temporal action localization (PTAL) is proposed as a reliable solution, as it requires only one labeled frame per action instance without precise boundaries.
As a pioneering method, SF-Net~\cite{SF-Net} employs an iterative strategy to mine pseudo action and background frames, thereby offering supplementary supervision throughout the training process.
Subsequent research extends this paradigm by exploring effective strategies for leveraging such limited supervision.
Typically, LACP~\cite{LACP} employs a greedy algorithm to generate a densely optimal sequence and leverages point annotations to learn the completeness of action instances.
CRRC-Net~\cite{crrc-net} is  proposed to further tackle intra-action variation and pseudo-label noise via co-supervised feature learning and probabilistic label mining.
Recently, SNPR~\cite{snpr} proposes two neighbor-guided strategies to improve training efficiency and reduce label noise caused by sparse annotations and unreliable predictions.

Despite progress, the usage of point annotations of most existing methods remains limited to supervising snippet-level classification.
Specifically, in a training process, an action locator relies on point annotations to supervise action classification at the snippet level and continuously mines pseudo-labels to enrich the supervisory information, as illustrated in \textbf{Task 1} of Figure~\ref{fig:insight}.
\textcolor{black}{
\textcolor{black}{However, these methods do not explicitly model the temporal relationships within actions, although such relationships are crucial for understanding its temporal structure and localizing complete actions.}
To address this limitation, we introduce temporal consistency learning under point supervision, which aims to help the model capture temporal structures shared by action instances of the same category across different videos.
In this work, such temporal consistency is considered from three aspects: temporal context, temporal order, and action regularity.
Accordingly, we design three self-supervised auxiliary tasks based on point annotations, as shown in \textbf{Tasks 2, 3, and 4} of Figure~\ref{fig:insight}.
Action Completion focuses on temporal context modeling, Action Order Understanding promotes temporal order understanding, and Action Regularity Understanding encourages action regularity learning.}
The three point-based proxy tasks are described as follows:

(i) \textbf{Action Completion (Figure~\ref{fig:insight} Task 2):} predicting the masked features of an annotated action snippet based on preceding and succeeding frames, thereby enabling the action locator to better model the temporal context of the action.

(ii) \textbf{Action Order Understanding (Figure~\ref{fig:insight} Task 3):} inferring whether an action sequence is temporally reversed, to strengthen the model's awareness of action order.

(iii) \textbf{Action Regularity Understanding (Figure~\ref{fig:insight} Task 4):} determining whether the action sequence is timely continuous, thus improving the model's understanding of action regularities.

\textcolor{black}{These three auxiliary tasks are jointly optimized with the main snippet classification task in a unified multi-task learning framework, enabling the shared representation to capture richer temporal structures for action localization.}
\textcolor{black}{By extracting richer temporal information from sparse point supervision, our method further improves the trade-off between annotation efficiency and localization performance in PTAL. 
Moreover, the proposed framework only introduces compact auxiliary branches during training, which are removed during inference, \textcolor{black}{without introducing additional inference overhead or requiring substantial architectural redesign.}}
Extensive experiments on four widely used datasets demonstrate the effectiveness of our method.

Our contributions are summarized as follows:
\begin{compactitem}
\item \textcolor{black}{We analyze the lack of explicit temporal consistency modeling objectives in current point-supervised methods and explore the benefits of temporal consistency modeling for action localization.}

\item \textcolor{black}{ We introduce a unified multi-task temporal consistency learning framework by integrating three point-guided temporal proxy tasks, namely action completion, action order understanding, and action regularity understanding, into the sparse point-supervised TAL setting. These tasks encourage the model to learn temporal context, temporal order, and action regularity, thereby facilitating temporal consistency learning.}

\item 
Extensive experimental results on four widely used datasets: THUMOS’14, GTEA, BEOID, and ActivityNet 1.3, show the effectiveness of the proposed method 
compared to several state-of-the-art methods.
\end{compactitem}

\section{Related Work}

\subsection{Temporal Action Localization}
\noindent\textbf{Fully-Supervised Temporal Action Localization} relies on accurate annotations of action start and end timestamps, assigning precise action labels to every snippet in the training stage. 
This precise labeling leads to strong performance in both action classification and localization. Current approaches primarily include:
(1) Proposal-based methods~\cite{BSN,BMN}, which generate temporal windows indicating action boundaries and refine these boundaries using regression modules.
(2) Proposal-free methods~\cite{actionformer}, directly predict the probabilities for each snippet, with the snippet of highest probability selected as the action boundary proposal.
Despite their excellent performance, these methods still face significant challenges due to the high cost of detailed labeling for each video snippet.

\noindent\textbf{Weakly-Supervised Temporal Action Localization} only provides video-level labels to localize action instances.
Some studies adopt the multiple instance learning (MIL) mechanism to aggregate top-k snippet-level predictions into a video-level classification result.
Some other works apply attention-based methods to boost detection performance.
For example, UntrimmedNets~\cite{wang2017untrimmednets} employs a soft attention mechanism to effectively retrieve relevant video segments.
HAM-Net~\cite{HAM-Net} proposes a hybrid attention mechanism encoding action completeness, while ASCN~\cite{ASCN} similarly adopts hybrid attention integrated with frame differences to capture local contextual features.
However, due to the unavailability of frame-level annotations, the performance of these models is significantly lower compared to fully-supervised methods.

\noindent\textbf{Point-Level Weakly-Supervised Temporal Action Localization} differs from traditional video-level Weakly-Supervised Temporal Action Localization in that it annotates a few numbers of action frames, \textit{i.e.}, labeling only one frame per action instance.
The majority of approaches utilize point supervision to train action classifiers while also mining further reliable pseudo-labels to enhance supervisory information.
As a typical example, LACP~\cite{LACP} generates pseudo-background annotations and jointly searches for the optimal sequence with point annotations to facilitate completeness learning.
CRRC-Net~\cite{crrc-net} constructs and iteratively updates class-specific prototypes to achieve more reliable classification.
To obtain more precise action instances, several studies perform additional boundary refinement on classification-generated proposals.
TSPNet~\cite{TSPNet} leverages the human prior of annotating instance center regions to align proposal quality with confidence.
HR-Pro~\cite{HR-Pro} first learns discriminative snippet-level scores to generate reliable proposals, which are then refined using instance-level feature learning.
However, existing methods primarily focus on snippet-level classification, neglecting the temporal consistency of action instances. 
Therefore, in this paper, we introduce three point-based self-supervised temporal learning tasks aimed at modeling the temporal consistency inherent in action instances.

\subsection{Multi-Task Learning}
%
Multi-task learning~\cite{mtl_hard1, mtl_hard2, mtl_soft1, mtl_soft2} enhances task performance and model generalization through the joint learning of several related tasks, leveraging interaction and information sharing among them.
According to the degree of parameter sharing, multi-task learning methods can be broadly divided into two categories: hard parameter sharing and soft parameter sharing.
In the hard parameter sharing strategy~\cite{mtl_hard1, mtl_hard2}, multiple tasks share the parameters in the encoder, while each task maintains an independent decoder to generate task-specific outputs.
Unlike hard sharing, the soft parameter~\cite{mtl_soft1, mtl_soft2} uses separate encoders for each task and shares information through task-interaction modules.
In this paper, we adopt hard parameter sharing to jointly train the main task and self-supervised tasks, enhancing the model’s understanding of temporal consistency and its generalization.


\begin{figure*}[!t]
\centering
\includegraphics[width=\linewidth]{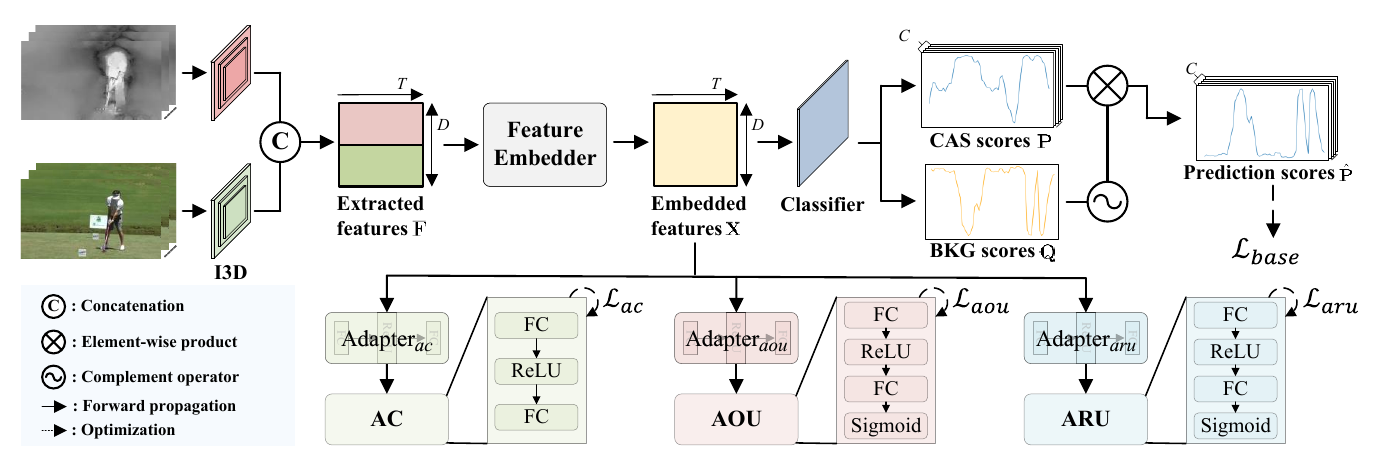}
\caption{The workflow of the proposed method.
The RGB and optical flow snippets of the input video are fed into the pretrained feature extractor to generate features $\mathbf{F}$, 
and are further embedded as $\mathbf{X}$.
Besides the conventional snippet classification task, the embedded features $\mathbf{X}$ are also used for joint training of the newly designed three temporal consistency tasks:  Action Completion (AC), Action Order Understanding (AOU), and Action Regularity Understanding (ARU).
}
\label{fig:method}
\end{figure*}

\section{Methodology}
In this section, we first give the problem definition of PTAL, our baseline architecture, as well as the baseline optimization loss in Sec.~\ref{sec: Preliminaries}.
Then, we introduce our multi-task learning setting in Sec.~\ref{sec: Multi-Task Learning Setting}.
Afterward, we present the details of our proposed three point-based self-supervised proxy tasks in Sec.~\ref{sec: Self-Supervised Learning Proxy Tasks}.
At last, we state the training and inference process in Sec.~\ref{sec: Training and Inference}.

\subsection{Preliminaries}~\label{sec: Preliminaries}

\noindent\textbf{Problem Definition.}
In the point-level weakly-supervised temporal action localization setting, each action instance in the given training video is weakly annotated with a single timestamp $t$ and its action category $y$, without providing temporal boundary information.
The goal of this task is to learn an action locator from the sparse annotations $P$, which predict the instance $(s, e, \hat{y}, c)$ of the input untrimmed video, where $s$ and $e$ indicate the start and end times of each action instance, $\hat{y}$ denotes the predicted action category, and $c$ represents the confidence score.

\noindent\textbf{Baseline Architecture.}
The input video is initially split into $T$ segments, each consisting of 16 frames.
Next, following the previous approach~\cite{SF-Net, LACP}, we employ a pre-trained I3D model to extract RGB and optical flow features from each snippet, which are then concatenated along the channel dimension.
The stacked snippet features can be represented $\mathbf{F}\in{\mathbb R}^{T\times D}$, where $D$ is the dimension of each snippet-level feature.
The extracted features are then fed into a feature embedding module to generate the embedded features ${\mathbf{X}}\in{\mathbb R}^{T\times D}$:
\begin{equation}
    \mathbf{X} = {Embed}(\mathbf{F})
\end{equation}
The feature embedder includes two transformer encoder layers, followed by a 1D convolutional layer and ReLU activation.
Afterward, $\mathbf{X}$ is fed into a snippet-level classifier to generate class-specific activation sequence (CAS) scores $\mathbf{P} \in \mathbb{R}^{T\times C}$ and the class-agnostic background scores $\mathbf{Q} \in \mathbb{R}^{T}$, where $C$ is the number of action categories and the classifier is composed of a 1D convolutional layer
with the Sigmoid function.
The final snippet-level predictions $\hat{\mathbf{P}}$ can be computed by element-wise production:
\begin{equation}
    \hat{\mathbf{P}} = \mathbf{P}\cdot ({1}-\mathbf{Q})
\end{equation}
\noindent\textbf{Baseline Optimization Loss.}
Since each action instance is labeled one point annotation and adjacent point annotations belong to separate action instances, we define pseudo-action snippets $\mathcal{T}^+ = \{t_i \}_{i=1}^{N_{act}}$ and pseudo-background snippets $\mathcal{T}^- = \{t_j \}_{j=1}^{N_{bkg}}$ using point annotations and the class-agnostic background scores, where $N_{act}$ and $N_{bkg}$ are the number of pseudo action snippets and background snippets, respectively. 
In specific, snippets located near a point annotation with lower class-agnostic background scores than a specified threshold 0.1 are classified as pseudo-action snippets, which share the same action category as the point annotation. In contrast, snippets between two adjacent point annotations that either have the highest class-agnostic background score or exceed a given threshold of 0.95 are classified as pseudo-background snippets. 
The pseudo-action snippets are used for foreground supervision, formulated as:
\begin{equation}
    {\mathcal L}_{cls}^{act}=\frac{1}{N_{act}} \sum_{c=1}^{C} \sum_{t\in{\mathcal{T}^+}}^{N_{act}}FL(\mathbf{P}_{t,c})
\end{equation}
Alternatively, pseudo-background snippets are used for background supervision:
\begin{equation}
    {\mathcal L}_{cls}^{bkg}= \frac{1}{N_{bkg}} \sum_{t\in{\mathcal{T}^-}}^{N_{bkg}} FL(\mathbf{Q}_{t})
\end{equation}
\textit{FL} represents the focal loss function~\cite{focalloss} as the same in~\cite{LACP}.
In addition, a feature contrastive loss~\cite{prototype_contra_loss} is adopted to separate actions and background in the feature space.
It can be formulated as:
\begin{equation}
\begin{aligned}
{\mathcal L}_{contra} = -\frac{1}{C}\sum_{c=1}^{C}\sum_{t\in \mathcal{T}^{+}_c}log(
                        \frac{exp(\bar{\mathbf{X}}_{t}\cdot\bar{\mathbf{m}}_c /\tau)}
                        {exp(\bar{\mathbf{X}}_{t}\cdot\bar{\mathbf{m}}_c/\tau)+\sum_{\forall k \neq c}exp(\bar{\mathbf{X}}_{t}\cdot\bar{\mathbf{m}}_k/\tau)} \\
                        +\frac{exp(\bar{\mathbf{X}}_{t}\cdot\bar{\mathbf{m}}_c/\tau)}
                        {exp(\bar{\mathbf{X}}_{t}\cdot\bar{\mathbf{m}}_c/\tau)+\sum_{\forall t_j \in \bar{\mathcal{T}}^-}exp(\bar{\mathbf{X}}_{t_j}\cdot\bar{\mathbf{m}}_c/\tau)})
\end{aligned}
\end{equation}
where $\mathbf{m}_c$ represents the prototype vector of the $c$-th action class, obtained by mean pooling the features of the $c$-th action snippets.
$\tau$ denotes the temperature parameter, ${\bar{\mathbf{X}} }$ denotes the L2 normalized features of $\mathbf{X}$.

In total, the baseline optimization loss can be expressed as:
\begin{equation}
    \mathcal{L}_{base} = \lambda_{1} \mathcal{L}_{cls}^{act} + \lambda_{2}{\mathcal L}_{cls}^{bkg} + \lambda_{3}{\mathcal L}_{contra}
\end{equation}
where $\lambda_{1}$, $\lambda_{2}$, and $\lambda_{3}$ are the weighting parameters for $\mathcal{L}_{cls}^{act}$, ${\mathcal L}_{cls}^{bkg}$, and ${\mathcal L}_{contra}$.
\subsection{Multi-Task Learning Setting}~\label{sec: Multi-Task Learning Setting}
To boost the temporal consistency of actions, we introduce three auxiliary tasks into the baseline, forming a unified multi-task learning framework.
The overall multi-task learning architecture is presented in Figure~\ref{fig:method}.
Serving as shared features, the embedded features $\textbf{X}$ are input to three task-specific adapters, each projecting them into a unique feature space:
\begin{equation}
    \textbf{X}^{ac} = \text{Adapter}_{{ac}}(\textbf{X})
\end{equation}
\begin{equation}
    \textbf{X}^{aou} = \text{Adapter}_{{aou}}(\textbf{X})
\end{equation}
\begin{equation}
    \textbf{X}^{aru} = \text{Adapter}_{{aru}}(\textbf{X})
\end{equation}
Each adapter consists of a fully connected (FC) layer, a ReLU activation function, and another fully connected layer.
Subsequently, the three task-specific features are fed into their corresponding task heads.

\subsection{Point-based Self-Supervised Learning}~\label{sec: Self-Supervised Learning Proxy Tasks}
We define the action localization branch in the upper part of Figure~\ref{fig:method} as the main task (\textbf{Task 1}), while the three auxiliary task branches in the lower part of Figure~\ref{fig:method} are denoted as \textbf{Tasks 2, 3, and 4}, respectively.

\noindent\textbf{Task2: Action Completion.}
To improve the model’s temporal context awareness, we design a task inspired by BERT’s masked language model mechanism~\cite{bert} in NLP, where the model learns to predict the feature representation of a labeled middle snippet based on its context.
The labeled snippet is termed as $\textbf{X}_i^{ac}$, and $\textbf{X}^{ac}=\{\textbf{X}_i^{ac}\}_{i=1}^T$ represents all $T$ action completion task-specific features.
We input the forward-left snippets $\overset{\rightarrow}{\mathbf{S}}{}_{i}^{ac} = [\textbf{X}_{i-t}^{ac}, \dots, \textbf{X}_{i-1}^{ac}]$ and the backward-right snippets $\overset{\leftarrow}{\mathbf{S}}{}_{i}^{ac} = [\textbf{X}_{i+t}^{ac}, \dots, \textbf{X}_{i+1}^{ac}]$
into our action completion task head $\text{H}_{ac}$:
\begin{equation}
    \hat{\textbf{X}}_{i}^{ac} = \text{H}_{ac}([\overset{\rightarrow}{\mathbf{S}}{}_{i}^{ac}, \overset{\leftarrow}{\mathbf{S}}{}_{i}^{ac}])
\end{equation}
where $\hat{\textbf{X}}_{i}^{ac}$ represents the predicted labeled snippet feature, and $\text{H}_{ac}$ includes a fully connected layer, followed by a ReLU activation function and a subsequent fully connected layer.
We calculate the cosine distance to evaluate the semantic gap between the predicted results $\hat{\textbf{X}}_{i}^{ac}$ and the ground truth ${\textbf{X}}_{i}^{ac}$:
\begin{equation}
    \mathcal{L}_{ac} = 1-\frac{1}{M} \sum_{j=1}^{M}(\frac{{\textbf{X}}_{j}^{ac} \cdot \hat{\textbf{X}}_{j}^{ac}}{\|{\textbf{X}}_{j}^{ac}\|_{2} \cdot \|\hat{\textbf{X}}_{j}^{ac}\|_{2}})
\end{equation}
where $M$ is the number of point-labels, $\|\cdot\|_{2}$ represents the L2 normalized operation.
The objective of this task is to minimize $\mathcal{L}_{ac}$.

\noindent\textbf{Task3: Action Order Understanding.}
Learning the temporal action sequence order is effective in modeling temporal consistency of action instances.
The goal of this task is to distinguish between forward and backward sequences.
It motivates the model to capture more discriminative information about the order of temporal action sequences.
We define the original forward sequence $\overset{\rightarrow}{\mathbf{S}}{}_{i}^{aou} = [\textbf{X}_{i-t}^{aou}, \dots, \textbf{X}_{i}^{aou}, \dots, \textbf{X}_{i+t}^{aou}]$ as a positive sample labeled 1, and its reversed version $\overset{\leftarrow}{\mathbf{S}}{}_{i}^{aou} = [\textbf{X}_{i+t}^{aou}, \dots, \textbf{X}_{i}^{aou}, \dots, \textbf{X}_{i-t}^{aou}]$ as a negative sample labeled 0.
Next, both $\overset{\rightarrow}{\mathbf{S}}{}_{i}^{aou}$ and $\overset{\leftarrow}{\mathbf{S}}{}_{i}^{aou}$ are passed to the action order discriminator $D_{aou}$:
\begin{equation}
    \overset{\rightarrow}{pred} = D_{aou}(\overset{\rightarrow}{\mathbf{S}}{}_{i}^{aou})
\end{equation}

\begin{equation}
    \overset{\leftarrow}{pred} = D_{aou}(\overset{\leftarrow}{\mathbf{S}}{}_{i}^{aou})
\end{equation}
where the $\overset{\rightarrow}{pred}$ and $\overset{\leftarrow}{pred}$ represent the predicted results of $\overset{\rightarrow}{\mathbf{S}}{}_{i}^{aou}$ and $\overset{\leftarrow}{\mathbf{S}}{}_{i}^{aou}$, respectively.
This order discriminator $D_{aou}$ architecture includes one fully connected layer, followed by a ReLU activation, another fully connected layer, and a Sigmoid function as the output.
To encourage the model to capture the ordering of actions, we minimize the following loss function:
\begin{equation}
    \mathcal{L}_{aou} = -\frac{1}{M} \sum_{j=1}^{M}[\log{\overset{\rightarrow}{pred}_j} + \log(1-{\overset{\leftarrow}{{pred}_j}})]
\end{equation}

\noindent\textbf{Task4: Action Regularity Understanding.}
Understanding the regularity of actions is crucial for temporal action detection.
The purpose of this task is to distinguish between regular and irregular action sequences.
It drives the model to focus on identifying key cues that reflect the regularity of temporal action sequences.
\textcolor{black}{
We define the features of regular action segments as positive samples with label 1, denoted as
\(\bar{\mathbf{S}}_{i}^{aru} = [\mathbf{X}_{i-t}^{aru}, \dots, \mathbf{X}_{i}^{aru}, \dots, \mathbf{X}_{i+t}^{aru}]\).
To construct negative samples with label 0, we randomly shuffle the elements in \(\bar{\mathbf{S}}_{i}^{aru}\) to obtain an irregular sequence:
\(\tilde{\mathbf{S}}_{i}^{aru} = \operatorname{Shuffle}(\bar{\mathbf{S}}_{i}^{aru})\).
Here, \(\operatorname{Shuffle}(\cdot)\) denotes a random permutation of the snippets in the sequence, with the constraint that the generated sequence is neither the original sequence nor its reversed version.
During training, the irregular sequence is randomly sampled rather than manually fixed. If the sampled sequence corresponds to the reversed sequence, we re-shuffle it until a non-reversed irregular sequence is obtained.
This constraint avoids overlap with the AOU task, since AOU focuses on distinguishing forward and reversed temporal structures, whereas ARU aims to distinguish regular and irregular temporal patterns.}
We then feed both $\overset{-}{\mathbf{S}}{}_{i}^{aru}$ and $\overset{\sim}{\mathbf{S}}{}_{i}^{aru}$ into the action regularity discriminator $D_{aru}$ to generate prediction scores $\overset{-}{pred}$ and $\overset{\sim}{pred}$:
\begin{equation}
    \overset{-}{pred} = D_{aru}(\overset{-}{\mathbf{S}}{}_{i}^{aru})
\end{equation}

\begin{equation}
    \overset{\sim}{pred} = D_{aru}(\overset{\sim}{\mathbf{S}}{}_{i}^{aru})
\end{equation}
$D_{aru}$ shares the same architecture as $D_{aou}$.
We minimize the following loss function to help the model capture the inherent regularity of actions:
\begin{equation}
    \mathcal{L}_{aru} = -\frac{1}{M} \sum_{j=1}^{M}[\log{\overset{-}{pred}_j} + \log(1-{\overset{\sim}{{pred}_j}})]
\end{equation}

\subsection{Training and Inference}~\label{sec: Training and Inference}

\noindent\textbf{Training.} The overall objective loss function is formulated as follows:
\begin{equation}
    \mathcal{L}_{total} = \mathcal{L}_{base} + \lambda_{ac}\mathcal{L}_{ac} + \lambda_{aou}\mathcal{L}_{aou} + \lambda_{aru}\mathcal{L}_{aru}
\end{equation}
where $\lambda_{ac}$, $\lambda_{aou}$, and $\lambda_{aru}$ are hyperparameters that control the trade-off between the auxiliary tasks.


\noindent\textbf{Inference.}
\textcolor{black}{
During inference, we retain only the main branch trained with multi-task learning as the detector, without relying on the auxiliary task modules, to generate final prediction scores.
Specifically, we first apply a video-level threshold \(\theta_{v}\) to determine which action categories are present in the input video. 
The video-level score for each category is computed by averaging the snippet-level CAS scores ${\mathbf{P}}$ that are among the top \(k_{act}\) highest for that category.
For the selected categories, we then apply the snippet-level threshold \(\theta_{p}\) to the snippet-level final scores $\hat{\mathbf{P}}$ to select candidate action snippets.
Consecutive candidate snippets are merged into a single temporal proposal, and the start and end times of the proposal are determined by the first and last snippets in the merged sequence.
The predicted category \(\hat{y}\) is given by the selected action class, and the confidence score \(c\) of each proposal is computed using the outer-inner-contrast score~\cite{shou2018autoloc}.
Finally, \textcolor{black}{Soft-NMS is applied to suppress redundant, highly overlapping proposals} and produce the final localization results \((s,e,\hat{y},c)\).
}

\section{Experiments}

\subsection{Datasets and Evaluation}
We conduct experiments on the four commonly-used temporal action localization datasets, including THUMOS'14~\cite{th14}, BEOID~\cite{beoid}, GTEA~\cite{gtea}, and ActivityNet 1.3~\cite{activitynet}.

\noindent\textbf{THUMOS'14}~\cite{th14} serves as a challenging dataset for Temporal Action Localization, containing 413 untrimmed sports videos across 20 action categories. The average number of action instances per video is 15. 
The dataset is split into 200 videos for training and 213 videos for testing, as per the standard division.

\noindent\textbf{BEOID}~\cite{beoid} offers 58 video samples categorized into 34 operation classes across 6 different locations. It consists of 46 videos for training and 12 for testing.

\noindent\textbf{GTEA}~\cite{gtea} consists of 28 untrimmed videos capturing 7 fine-grained daily activities in a kitchen setting. We follow the standard split, using 21 videos for training and 7 for testing.

\noindent\textbf{ActivityNet 1.3}~\cite{activitynet} includes 10,024 training videos, 4,926 validation videos, and 5,044 test videos, covering 200 action categories. On average, each video contains 1.6 action instances.

\noindent\textbf{Evaluation Metrics}.
For a fair comparison, we evaluate the action localization performance on the four datasets using mean Average Precision (mAP) at various Intersection-over-Union (IoU) thresholds. 
If the generated proposal’s IoU exceeds the set IoU threshold and matches the correct category, it is regarded as a positive sample.

\subsection{Implementation Details}
To make a fair comparison, we follow the previous methods to split each video into non-overlapping 16-frame snippets and use a two-stream I3D~\cite{i3d} network pre-trained on Kinetics-400 as the feature extractor.
The input feature dimension $D$ is 2048, the adapter feature dimension is 512.
\textcolor{black}{The feature embedder consists of a Transformer encoder with a hidden dimension of 512 and 8 attention heads, followed by a 1D convolution layer with kernel size 3 and stride 1. The model is trained for 1000 iterations with a batch size of 16 using the Adam optimizer, with a learning rate of $1e{-4}$ and a weight decay of $1e{-3}$.}
Empirically, $\lambda_1$, $\lambda_2$, $\lambda_3$, $\lambda_{ac}$, $\lambda_{aou}$, and $\lambda_{aru}$ are 0.5, 1, 1, 0.5, 0.5, and 0.5, respectively.
\textcolor{black}{We set the video-level threshold \(\theta_v\) to 0.5 and vary the snippet-level threshold \(\theta_p\) from 0 to 0.25 with a step size of 0.05.}
The temperature parameter $\tau$ is 0.1, and each temporal window (\textit{i.e.}, the temporal sequence composed of the labeled snippet and its surrounding snippets) includes 5 snippet features.
Our entire system is implemented using PyTorch, and all experiments are conducted on a single RTX-3090 GPU.

\begin{table*}[t]
\centering
\caption{
Comparison with the state-of-the-art methods on THUMOS'14.
We further compare our method with both fully-supervised and weakly-supervised approaches.
For fairness, we adopt the same point annotations as used in~\cite{LACP}.
Moreover, to enable a fair comparison with proposal-based two-stage methods~\cite{TSPNet, HR-Pro, dcm}, we also apply a proposal-based boundary adjustment similar to~\cite{HR-Pro}.
Results marked with $^{\dagger}$ indicate our method with proposal-based refinement.} 
\resizebox{\textwidth}{!}{
\begin{tabular}{c|l|ccccccc|ccc}
\hline\hline

\multirow{2}{*}{Supervision}       & \multicolumn{1}{c|}{\multirow{2}{*}{Method}} &
\multicolumn{7}{c|}{mAP@IoU (\%)}  & AVG  & AVG  & AVG  \\
    & \multicolumn{1}{c|}{}  & 0.1  & 0.2  & 0.3  & 0.4  & 0.5  & 0.6  & 0.7 & (0.1:0.5)  & (0.3:0.7) & (0.1:0.7)    \\
    \hline\hline
    
\multirow{5}{*}{\begin{tabular}{c}Full \\supervised\end{tabular}}
    & TCANet \small {(CVPR'21)}  & -  & -  & 60.6  & 53.2  & 44.6  & 36.8  & 26.7  & -  & 44.3 & -\\
    & BMN+CSA \small {(ICCV'21)} & -  & -  & 64.4  &58.0  & 49.2 & 38.2  & 27.8  & -  & 47.5 & -\\
     & GCM \small {(TPAMI'21)}  & 72.5  & 70.9  & 66.5  & 60.8  & 51.9  & -  & -  & 64.5  & - & -\\
     & VSGN \small {(ICCV'21)} & -  & -  & 66.7  &60.4  & 52.4 & 41.0  & 30.4  & -  & 50.2 & -\\
    & AFSD \small {(CVPR'21)} & -  & -  & 67.3  & 62.4  & 55.5  & 43.7  & 31.1  & -  & 52.0 & -\\
    \hline\hline
    
\multirow{6}{*}{\begin{tabular}{c}Weakly \\supervised\end{tabular}}
    & P-MIL \small {(CVPR'23)} & 71.8  & 67.5  & 58.9  &49.0  & 40.0 & 27.1  & 15.1  & 57.4  & 38.0 & 47.0\\
    & Zhou~\etal \small {(CVPR'23)}   & 74.0  & 69.4  & 60.7  & 51.8  & 42.7  & 26.2  & 13.1  & 59.7  &38.9  & 48.3\\
    & PivoTAL \small {(CVPR'23)} & 74.1  & 69.6  & 61.7  &52.1  & 42.8 & 30.6  & 16.7  & 60.1  & 40.8 & 49.6\\
    & ISSF \small{(AAAI'24)} &72.4 &66.9 &58.4 &49.7 &41.8 &25.5 &12.8 &57.8 &37.6 &46.8 \\
    &PVLR  \small{(MM'2024)} &74.9 &69.9 &61.4 &53.1 &45.1 &30.5 &17.1 &60.9&-&50.3 \\
    &NoCo  \small{(AAAI'2025)} &75.2 &70.7 &63.0 &54.1 &44.0 &31.7 &17.7 &61.4&42.1&50.9 \\

    \hline
    \multirow{16}{*}{\begin{tabular}{c}Point\\supervised\end{tabular}}
    &ARST~\cite{ARST} \small {(CVPR'19)} &24.3 &19.9 &15.9 &12.5 &9.0 &- &- &16.30 &- &- \\
    & SF-Net~\cite{SF-Net} \small {(ECCV'20)}  & 68.3  & 62.3  & 52.8  & 42.2  & 30.5  & 20.6  & 12.0   & 51.2  & 31.6  & 41.2\\
    & LACP~\cite{LACP} \small {(ICCV'21)}  & 75.7  & 71.4  & 64.6  & 56.5  & 45.3  & 34.5  &  21.8  & 62.7  &  44.5  & 52.8\\
    & BackTAL~\cite{backtal} \small {(TPAMI'21)} &-&-&54.4&45.5&36.3&26.2&14.8&-&35.4&- \\
    & CRRC-Net~\cite{crrc-net} \small{(TIP'22)} & 77.8  & 73.5  & 67.1  & 57.9  & 46.6  & 33.7  &  19.8  & 64.6  &  45.1  & 53.8\\
    & SMBD~\cite{SMBD} \small{(ECCV'24)} & -  & -  & 66.0  & 57.9  & 47.0  & 36.0  &  22.0  & 64.2  &  45.7  & -\\
    & SNPR~\cite{snpr} \small{(TIP'24)} & 77.9  & 73.9  & 66.6  & 59.4  & 48.6  & 36.7  &  22.7  & 65.3  &  46.8  & 55.1\\
    & AAPL~\cite{AAPL} \small{(AAAI'25)} & 64.3  & -  & 54.6  & -  & 35.2  & -  &  14.0  & -  &  -  & 42.8\\
    &  $\textbf{Ours}$
  & \textbf{82.5}  & \textbf{78.2}  & \textbf{71.5}  & \textbf{62.3}  & \textbf{51.4}  & \textbf{38.0}  & \textbf{23.3}  & \textbf{69.2}  &  \textbf{49.3} &\textbf{58.2} \\
    \cline{2-12}
    & DCM~\cite{dcm} \small {(ICCV'21)}  & 72.3  & 64.7  & 58.2  & 47.1  & 35.9  & 23.0  & 12.8   & 55.6  & 35.4  & 44.9\\
    &SF-Net+{TSPNet}~\cite{TSPNet} \small {(CVPR'24)} &75.5 &- &57.9 &- &33.1 &- &12.3 &-&-&44.8 \\
    &BackTAL+{TSPNet}~\cite{TSPNet} \small {(CVPR'24)} &77.1 &- &63.5 &-&43.4 &-&19.7 &-&-&51.7 \\
    &  TSPNet~\cite{TSPNet} \small {(CVPR'24)}  & 82.3  & 77.6  & 70.1  & 60.0  & 49.4  & 37.6  & 24.5  & 67.9  &  48.3 &57.4 \\
    &  HR-Pro~\cite{HR-Pro} \small {(AAAI'24)}  & 85.6  & 81.6  & 74.3  & 64.3  & 52.2  & 39.8  & 24.8  & 71.6  &  51.1 &60.3 \\
    &  QROT~\cite{QROT} \small {(CVPR'25)}  & -  & -  & 73.1  & 64.4  & 54.3  & 41.3  & 27.4  & 72.3  &  52.1 &- \\
    &  $\textbf{Ours}^{\dagger}$
  & \textbf{85.9}  & \textbf{82.3}  & \textbf{75.5}  & \textbf{66.5}  &\textbf{54.7}  & \textbf{41.6}  & \textbf{25.8}  & \textbf{73.0}  &  \textbf{52.8} &\textbf{61.8} \\
    
    \hline\hline
    
\end{tabular}
}
\label{table:thumos_benchmark}
\end{table*}

\subsection{Comparison with State-of-The-Art Methods}
We evaluate our model’s performance and compare it with the latest PTAL methods, as well as some fully-supervised and weakly-supervised methods.

\noindent\textbf{THUMOS'14.} In Table~\ref{table:thumos_benchmark}, our method achieves the highest average mAP (0.1:0.7) 58.2\% compared to all point-level and video-level weakly-supervised temporal action localization methods.
In comparison to previous one-stage PTAL methods, the average mAP (0.1:0.7) of our method rises from 3.1\% (Ours \textit{v.s.} SNPR) to 17\% (Ours \textit{v.s.} SF-Net).
Compared to the best-performing video-level weakly supervised method NoCo, our approach achieves a 7.3\% higher score on the average mAP for IoU thresholds of 0.1:0.7.
Our proposal-based refinement results also surpass other proposal-based PTAL methods, such as HR-Pro~(61.8\% \textit{v.s.} 60.3\%) and TSPNet (61.8\% \textit{v.s.} 57.4\%). On the average mAP for IoU thresholds of 0.3:0.7, our method even outperforms certain fully-supervised approaches, such as AFSD~(52.8\% \textit{v.s.} 52.0\%).
The above comprehensive comparison results demonstrate the effectiveness of our proposed method.

\noindent\textbf{GTEA \& BEOID \& ActivityNet 1.3.}
We further evaluate the effectiveness and robustness of the proposed method on three other action detection benchmark datasets in Table~\ref{table:multi_benchmark}.
%
On both the GTEA and BEOID datasets, our method achieves a notable performance gain, outperforming SMBD by 2.9\% and 2.8\% on the average mAP(0.1:0.7), respectively.
When compared to proposal-based methods, our proposal-level boundary adjustment surpasses HR-Pro by 4.0\% and 3.5\%.
On the large-scale dataset ActivityNet 1.3, our method also achieves competitive results, with an average mAP (0.5:0.95) of 26.3\%, only slightly behind SNPR’s 26.5\%.
In addition, our method achieves 27.4\% average mAP (0.5:0.95) via proposal refinement, outperforming HR-Pro (27.1\%) by 0.3\%.

\begin{table*}[!t]
\centering
\caption{
Comparisons of detection performance on GTEA, BEOID, and ActivityNet 1.3 datasets.
The symbol ${\dagger}$  indicates our results with boundary adjustment, following the same proposal-based refinement strategy as~\cite{HR-Pro}.
}
\resizebox{\textwidth}{!}{
\begin{tabular}{l|ccccc|ccccc|cccc}
\hline
\hline
 & \multicolumn{5}{c|}{GTEA} & \multicolumn{5}{c|}{BEOID} & \multicolumn{4}{c}{ActivityNet 1.3} \\
\hline
\multirow{2}{*}{Method}& \multicolumn{5}{c|}{mAP@IoU (\%)} & \multicolumn{5}{c|}{mAP@IoU (\%)} & \multicolumn{4}{c}{mAP@IoU (\%)} \\
\multicolumn{1}{c|}{} & 0.1  & 0.3  & 0.5  & 0.7 &AVG[0.1:0.7] & 0.1  & 0.3  & 0.5  & 0.7 &AVG[0.1:0.7] & 0.5  & 0.75  & 0.95 &AVG[0.5:0.95] \\
\hline 
SF-Net \small(ECCV'20) & 58.0  & 37.9  & 19.3  & 11.9  & 31.0 & 62.9  & 40.6  & 16.7  & 3.5  & 30.9 & - & - & - & - \\
LACP \small(ICCV'21) & 63.9  & 55.7  & 33.9  & \textbf{20.8}  & 43.5  & 76.9  & 61.4  & 42.7  & 25.1  & 51.8 & 40.4  & 24.6  & 5.7 & 25.1 \\
CRRC-Net \small(TIP'22)  & -  & -  & - & - & - & -  & -  & - & - & - & 39.8  & 24.1  & 5.9 & 24.0 \\
SMBD \small {(ECCV'24)}  & 75.0  & 61.3   & 41.1  & 14.2 & 47.4  & \textbf{78.2}  & 71.0   & 52.5  & 25.2 & 57.4 &- &- &- &- \\
SNPR \small {(TIP'24)}  & 74.3 & 62.8   & 35.7  & 13.7 & 46.6 & 77.2  & 64.3   & 44.0  & 24.5 & 53.1 
&41.3 &\textbf{30.9} &4.8 &\textbf{26.5} \\
AAPL \small(AAAI'25)  & 70.3  & 54.4  & 37.7 & 23.4 & 46.3 & 75.5  & 67.6  & 48.5 & 26.3 & 55.2 & 39.6  & 24.3  & \textbf{5.6} & 24.7 \\
\textbf{Ours} &\textbf{76.6} &\textbf{64.3} &\textbf{45.3} &{17.7} &\textbf{50.3}
&{77.1} &\textbf{72.1} &\textbf{55.7} &\textbf{28.4} &\textbf{60.2}
&\textbf{42.3} &{25.9} &{5.5} &{26.3} \\
\hline
DCM \small(ICCV'21)  & 59.7  & 38.3  & 21.9  & 18.1  & 33.7 & 63.2  & 46.8  & 20.9  & 5.8  & 34.9 & - & - & - & - \\
{HR-Pro} \small(AAAI'24)  & {72.6}  & {61.1}  & {37.3}  & 17.5  & {47.3}  
& {78.5}  & {72.1}  & {55.3}  & {26.1}  & {59.4}
& {42.8}  & \textbf{27.2}  & \textbf{8.0} & {27.1} \\
{TSPNet} \small(CVPR'24) & {74.6}  & {60.9}  & {39.5}  & {16.6}  & {49.0}  
& \textbf{83.8}  & {73.0}  & {51.1}  & {23.8}  & {59.6}
& {-}  & {-}  & {-} & {-} \\
\textbf{Ours}$^{\dagger}$ &\textbf{76.4} &\textbf{68.8} &\textbf{48.4} &\textbf{18.3} &\textbf{53.0}
&{79.5} &\textbf{75.7} &\textbf{62.5} &\textbf{29.6} &\textbf{63.1}
&\textbf{43.9} &{27.1} &{5.6} &\textbf{27.4} \\

\hline 
\hline

\end{tabular}
}
\label{table:multi_benchmark}
\end{table*}

\subsection{Evaluation of Temporal Consistency} 
\noindent\textbf{Proxy Task Evaluation.}
Besides evaluating temporal action localization performance, we also evaluate the performance of three self-supervised tasks to measure the model’s perception of temporal consistency in Table~\ref{table:proxy_task_results}.
In detail, we perform Gaussian sampling on the action instances from the THUMOS'14 test set, process the sampled points as described in Section~\ref{sec: Self-Supervised Learning Proxy Tasks}, and then input them into three task heads for performance evaluation.
In the Action Completion task, the cosine similarity between the predicted snippet features and the corresponding target snippet is 0.992, and the cosine distance is 0.008, highlighting the excellent ability to predict action snippets using temporal contextual information.
In the Action Order Understanding task, our order discriminator $D_{aou}$ achieves 92.9\% in AUC (Area Under the ROC Curve) for binary classification and a classification accuracy (ACC) of 85.7\%, which demonstrates the model’s strong understanding of temporal action sequence order.
In Action Regularity Understanding, the regularity discriminator also performed well in identifying action regularity, with an AUC of 85.9\% and an accuracy of 77.4\%.
This also indirectly indicates that the recognition of action regularity is inherently more challenging than action order prediction.

\begin{table}[H]
\centering
\caption{Quantitative results of different proxy tasks on the test split of THUMOS'14.
}
\resizebox{0.8\textwidth}{!}{

\begin{tabular}{c|c|c|c|c|c|c}
    \hline
    \textbf{Proxy Task} & 
    \multicolumn{2}{c|}{\textbf{AC}} & 
    \multicolumn{2}{c|}{\textbf{AOU}} & 
    \multicolumn{2}{c}{\textbf{ARU}} \\
    \hline
    \multirow{2}{*}{\textbf{Metric}} & Cos\_Sim & Cos\_Dist & AUC & ACC & AUC & ACC \\
    \cline{2-7}
                                     & 0.992 & 0.008 & 92.9\% & 85.7\% & 85.9\% & 77.4\% \\
    \hline
\end{tabular}

}
\label{table:proxy_task_results}
\end{table}

\subsection{Ablation Study}
We perform a set of ablation experiments in this section on THUMOS'14, as it is the most challenging dataset.


\noindent\textbf{Contribution of each component.}
In Table~\ref{table:ablation_studies}, we display the
action detection results of different combinations of proxy tasks.
``AC'', ``AOU'', and ``ARU'' represent our designed action completion, action order understanding, and action regularity understanding self-supervised tasks.
Regarding average mAP (0.1:0.7), the baseline (Row.~1) reaches a score of only 56.2\% without any proxy tasks.
In the setup with only one auxiliary task, we notice that ``ARU'' contributes the most~(Row.~4 \textit{v.s.} Row.~1), then the ``AOU''~(Row.~3 \textit{v.s.} Row.~1), and last the ``AC''~(Row.~2 \textit{v.s.} Row.~1).
When the number of proxy tasks increases to 2, we observe that both ``AC+AOU'' and ``AC+ARU'' achieve the same average mAP (0.1:0.7) of 57.6\%, while ``AOU+ARU'' had a slightly lower mAP of 57.5\%.
With the complete set of auxiliary tasks, the performance reaches its peak, with average mAP scores of 58.2 (0.1:0.7), 69.2 (0.1:0.5), and 49.3 (0.3:0.7).

\begin{table}[H]
\centering
\caption{
Ablation study on THUMOS'14.
$\uparrow$ denotes the relative gain between our full model and baseline.
}
\resizebox{0.8\textwidth}{!}{
\begin{tabular}{c|ccc|ccc}
    \hline\hline
    \multirow{2}{*}{\parbox[c]{0.8cm}{\centering Row\\Index}} &
    \multicolumn{3}{c|}{{Setup}} 
    & \multicolumn{3}{c}{AVG mAP@IoU(\%)} \\
    \cline{2-7}
    &AC & AOU & ARU  
      & [0.1:0.5] & [0.3:0.7] & [0.1:0.7] \\
    \hline
     1& &   &     & 67.4 & 46.9     & 56.2  \\
    2&$\checkmark$ &   &     & 67.5 & 47.5    & 56.6   \\
    3& &   $\checkmark$&     & 68.1 & 47.2    & 56.8   \\
   4&  &   &     $\checkmark$& 68.0 & 47.8    & 56.9   \\
   5& $\checkmark$ &$\checkmark$  &    &68.6 & 48.5    & 57.6 \\
    6& &$\checkmark$  & $\checkmark$   &68.9 & 48.1    & 57.5 \\
    7& $\checkmark$ &  & $\checkmark$   &68.4 & 49.0    & 57.6 \\
   8& $\checkmark$ &$\checkmark$   &$\checkmark$    &\textbf{69.2}{$^{\uparrow1.8}$} & \textbf{49.3}{$^{\uparrow2.4}$}    & \textbf{58.2}{$^{\uparrow2.0}$}  \\
    \hline\hline
\end{tabular}
}
\label{table:ablation_studies}
\end{table}

\noindent\textbf{Effectiveness of the length of temporal windows.}
In Table~\ref{table:ablation_studies_temporal_windows}, we evaluate the performance of our method using different temporal window lengths, each composed of the labeled snippet and its neighboring snippets.
Both longer and shorter temporal windows can degrade the performance.
This is because the longer temporal windows may introduce noise snippets (\textit{e.g.}, background segments visually similar to actions) and the shorter temporal windows cannot provide sufficient action information.
Therefore, we set the length of the temporal window to 5 as a balanced setting.

\begin{table}[H]
\centering
\caption{Ablation of different length of temporal windows on THUMOS'14.}
\resizebox{0.7\textwidth}{!}{
\begin{tabular}{c|cccc|c}
    \hline\hline
    {{Temporal}} & \multicolumn{4}{c|}{mAP@IoU (\%)}  & \textbf{AVG} \\
    {Windows} & 0.1  & 0.3  & 0.5  & 0.7 & [0.1:0.7] \\
    \hline
      {3}  &83.0 &71.4 &49.3 &23.0 &57.6  \\
      {5}  &82.5 &71.5 &51.4 &23.3 &58.2  \\
      {7}  &81.1 &70.3 &49.4 &22.3 &56.7  \\
    \hline\hline
\end{tabular}
}
\label{table:ablation_studies_temporal_windows}
\end{table}

\noindent\textbf{Influence of different feature extractors.}
We study the influence of different feature extractors on the performance of our method in Table~\ref{table:ablation_studies_feature}.
When utilizing the same type of pre-trained network (\textit{e.g.}, video swin transformer~\cite{video_swin}), we observe that a larger scale of pre-training data (Row.~4 \textit{v.s.} Row.~3) and more fine-grained convolutions (Row.~3 \textit{v.s.} Row.~2) can boost action detection performance.
I3D outperforms Video Swin because it utilizes not only RGB features but also optical flow features.
VideoMAEv2~\cite{videomaev2} obtains the best performance as it can learn strong feature representations from a vast amount of unlabeled video data, followed by further training on the K710.
To make a fair comparison with previous methods, I3D is selected as the feature extractor.

\begin{table}[H]
\centering
\caption{Ablation of different feature extractors on THUMOS'14.}
\resizebox{0.8\textwidth}{!}{
\begin{tabular}{c|c|cccc|c}
    \hline\hline
    \multirow{2}{*}{Feature Type} &\multirow{2}{*}{Pretrain} & \multicolumn{4}{c|}{mAP@IoU (\%)}  & \textbf{AVG} \\
    &  & 0.1  & 0.3  & 0.5  & 0.7 & [0.1:0.7] \\
    \hline
      {I3D (default)}&K400 & 82.5 & 71.5 & 51.4 & 23.3 & 58.2 \\
      {Video Swin}{$^{16\times7\times7}$} & K400&79.0 & 65.4 & 45.3 & 19.5 & 53.0 \\
      {Video Swin}{$^{8\times7\times7}$} & K400&79.8 & 67.4 & 46.3 & 19.7 & 54.0 \\
      {Video Swin}{$^{8\times7\times7}$} & K600&80.2 & 67.6 & 46.0 & 19.9 & 54.2 \\
      {VideoMAEv2} & K710&84.4 & 76.1 & 55.7 & 25.9 & 61.6 \\
    \hline\hline
\end{tabular}
}
\label{table:ablation_studies_feature}
\end{table}

\noindent\textbf{Impact of point annotations.}
We conduct an ablation study to explore the influence of point annotations from different distributions on THUMOS'14 in Table~\ref{table:ablation_studies_point}.
The performance of manual annotation and uniform annotation are degraded as some points fall near the boundary regions, leading to confusion between actions and background.
In contrast, center sampling performs better because it includes more salient action points, rather than the ambiguous points in the boundary regions.
Gaussian-based sampling achieves the best average mAP (0.1:0.7) of 58.2\ because it covers a wider range of diverse points.

\begin{table}[H]
\centering
\caption{Ablation of the point-level annotations from different distributions on THUMOS'14.}
\resizebox{0.8\textwidth}{!}{
\begin{tabular}{c|cccc|c}
    \hline\hline
    \multirow{2}{*}{Point Distribution} & \multicolumn{4}{c|}{mAP@IoU (\%)}  & \textbf{AVG} \\
      & 0.1  & 0.3  & 0.5  & 0.7 & [0.1:0.7] \\
    \hline
      {Manual} & 83.3 & 69.6 & 43.4 & 16.6 & 54.2 \\
      {Uniform} & 81.6 & 68.1 & 47.4 & 21.4 & 55.5 \\
      {Center} & 82.0 & 70.4 & 49.3 & 22.6 & 57.0 \\
      {Gaussian (default)} & 82.5 & 71.5 &51.4  & 23.3 & 58.2 \\
    \hline\hline
\end{tabular}
}
\label{table:ablation_studies_point}
\end{table}

\noindent\textbf{Generalization across different PTAL frameworks.}
\textcolor{black}{
To further evaluate the generalization ability of the proposed proxy tasks, we integrate them into three representative point-supervised TAL frameworks, including SF-Net, LACP, and HR-Pro$_{1}$. For a fair comparison, we keep the original framework of each method unchanged and only incorporate the proposed temporal proxy tasks as auxiliary supervision.
As shown in Table~\ref{table:Generalization}, the proposed proxy tasks consistently improve the localization performance of all three frameworks on THUMOS'14. Specifically, the average mAP over IoU thresholds from 0.1 to 0.7 increases from 40.2\% to 44.5\% for SF-Net, from 52.4\% to 54.7\% for LACP, and from 56.2\% to 58.2\% for HR-Pro$_{1}$. These consistent improvements indicate that the proposed proxy tasks are not limited to a specific baseline, but can provide effective temporal consistency supervision for different PTAL frameworks.
}
\begin{table}[H]
    \centering
    \caption{Results of incorporating our proxy tasks into different PTAL methods on THUMOS’14. $^*$ represents our reproduced results.}
    \resizebox{0.5\linewidth}{!}{
        \setlength{\tabcolsep}{1mm}{
            \begin{tabular}{l|cccc|c}
                \hline
                \multicolumn{1}{c|}{\multirow{2}{*}{Method}}  & \multicolumn{4}{c|}{mAP@IoU(\%)} & \textbf{AVG} \\
                \cline{2-6}
                  & 0.1 & 0.3 & 0.5 & 0.7 &[0.1:0.7] \\
                \hline
                SF-Net$^*$  & 68.2 & 52.5 & 29.1&10.4 &40.2 \\
                SF-Net + Ours  & 72.7 & 58.1 & 32.7&12.7&44.5 \\
                \hline
                LACP$^*$  & 75.3 & 64.3 & 45.1&20.6 &52.4 \\
                LACP + Ours  & 77.9 & 67.2 & 47.0&21.6&54.7 \\
                \hline
                HR-Pro$_{1}^*$  & 82.0 & 69.4 & 48.2&21.9 &56.2 \\
                HR-Pro$_{1}$ + Ours  & 82.5 & 71.5& 51.4&23.3 &58.2 \\
                \hline
            \end{tabular}            
        }
    }
     \label{table:Generalization}
\end{table}%

\noindent\textbf{Analysis of Auxiliary Loss Weights.}
\textcolor{black}{
We analyze the weighting strategy of the auxiliary losses. As shown in Table~\ref{table:ablation_studies}, the three temporal proxy tasks make comparable contributions to the final localization performance. When used individually, AC, AOU, and ARU improve the average mAP@[0.1:0.7] from 56.2\% to 56.6\%, 56.8\%, and 56.9\%, respectively. Their pairwise combinations also achieve similar performance, ranging from 57.5\% to 57.6\%. These results suggest that no single auxiliary task clearly dominates the others. Therefore, we use a shared weight for the three auxiliary losses, \textit{i.e.}, $\lambda_{ac}=\lambda_{aou}=\lambda_{aru}=\lambda$, to keep the framework simple and avoid excessive task-specific hyperparameter tuning.
We also conduct a sensitivity analysis of the shared weight $\lambda$, as reported in Table~\ref{tab:lambda_ablation}. The model achieves the best average mAP of 58.2\% when $\lambda=0.5$. A smaller $\lambda$ weakens the auxiliary supervision, while a larger $\lambda$ may overemphasize the auxiliary tasks and dilute the main localization objective. These results indicate that the shared-weight strategy is a reasonable and stable choice for balancing the auxiliary temporal proxy tasks and the main localization task.
}

\begin{table}[H]
\centering
\caption{Effect of coefficient $\lambda$ on average mAP@[0.1:0.7]. Here, $\lambda$ denotes the shared weight for the three auxiliary tasks: $\lambda_{\text{ac}} = \lambda_{\text{aou}} = \lambda_{\text{aru}}$.}
\resizebox{0.8\textwidth}{!}{%
\begin{tabular}{ccccccccccc}
\toprule
$\lambda$ & 0.1 & 0.2 & 0.3 & 0.4 & 0.5 & 0.6 & 0.7 & 0.8 & 0.9 & 1.0 \\
\midrule
mAP (\%) & 57.1 & 57.6 & 57.5 & 58.0 & \textbf{58.2} & 57.8 & 57.5 & 57.6 & 57.1 & 56.8 \\
\bottomrule
\end{tabular}%
}
\label{tab:lambda_ablation}
\end{table}

\noindent\textbf{Complementarity of AOU and ARU.}
\textcolor{black}{
AOU and ARU focus on different temporal properties: AOU distinguishes forward and reversed temporal structures, while ARU distinguishes regular and randomly shuffled irregular temporal patterns. To avoid overlap, reversed sequences are excluded when generating negative samples for ARU. As shown in Table~\ref{table:ablation_studies}, AOU and ARU individually improve the average mAP@[0.1:0.7] from 56.2\% to 56.8\% and 56.9\%, respectively. When combined, they further improve the performance to 57.5\%, outperforming either task alone. These results indicate that AOU and ARU provide complementary supervision and justify the necessity of distinguishing these two tasks.
}

\begin{figure}[H]
\centering
\includegraphics[width=\linewidth]{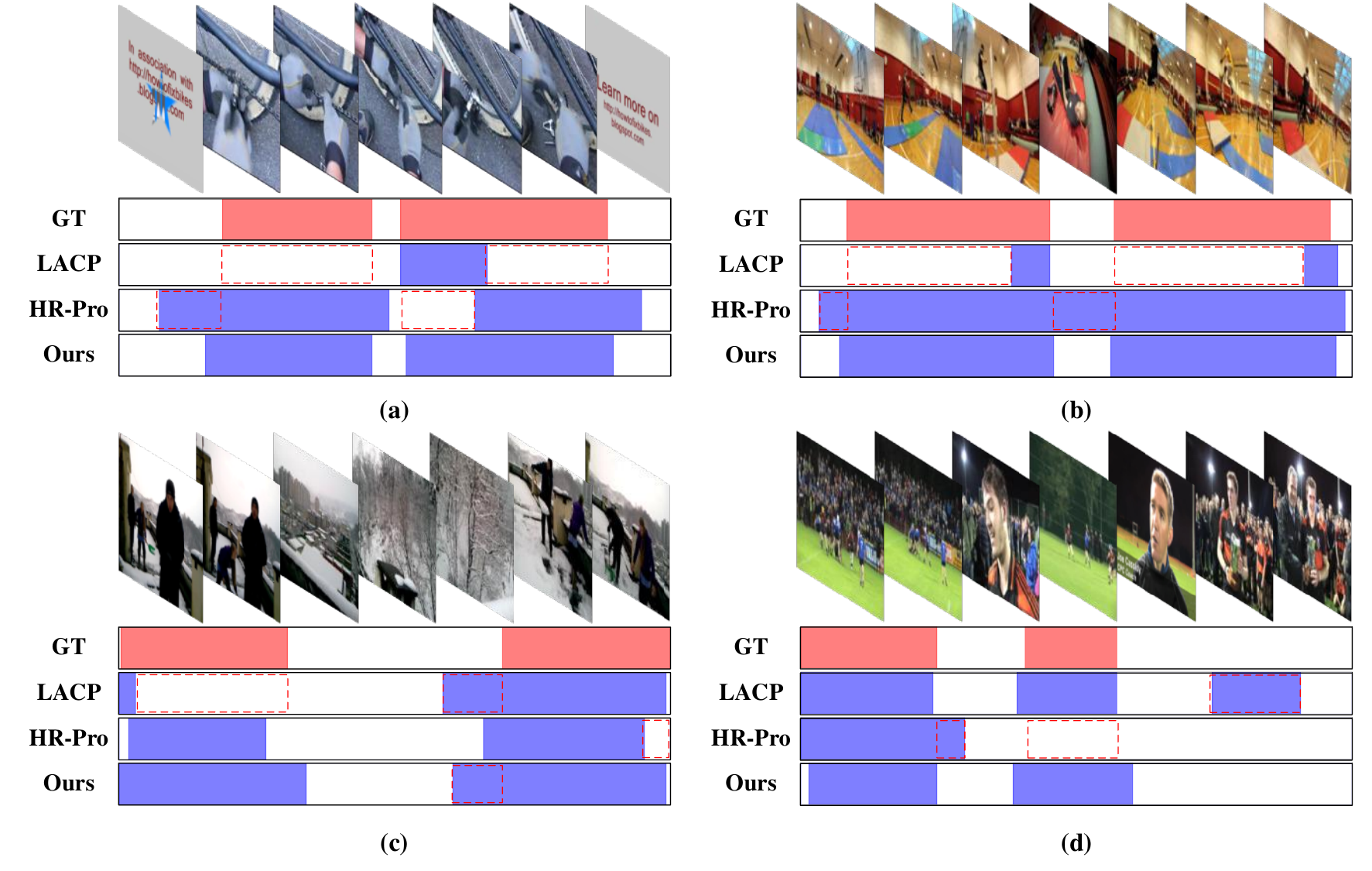}

\caption{Qualitative comparison between our proposed method, HR-Pro~\cite{HR-Pro}, and LACP~\cite{LACP} on ActivityNet 1.3.
We provide four cases of temporal action localization from different action classes, including ``Fixing bicycle'', ``Powerbocking'', ``Shoveling snow'', and ``Hurling''.
Prediction errors are highlighted with red dashed boxes.
The IoUs between our detection results and the ground truths are notably higher.
}
\label{fig:quality_result}

\end{figure}

\subsection{Qualitative Results}
\noindent\textbf{Qualitative comparison.}
In Figure~\ref{fig:quality_result}, we provide a qualitative comparison of our method against LACP and HR-Pro on the test videos of ActivityNet 1.3.
The results clearly show that our method generates more accurate action instance localization.
Taking Figure~\ref{fig:quality_result} (a) as an example, LACP predicts only one incomplete action instance, whereas the ground-truth annotation contains two complete action instances.
While HR-Pro detects two action instances, the left prediction includes extra segments, and the right one misses part of the action.
In contrast, our method not only successfully identifies all action instances but also accurately localizes their boundaries.
In Figure~\ref{fig:quality_result} (b), LACP identifies only a small portion of the action, while HR-Pro incorrectly merges two separate actions into one, ignoring the background segment in between. By contrast, our approach accurately captures two entire action sequences.
In Figure~\ref{fig:quality_result} (c), despite a slight over-detection in the second action instance, our method demonstrates superior overall completeness relative to the other two methods.
Figure~\ref{fig:quality_result} (d)  shows that the detection results of our method largely cover the ground truth, while LACP produces false positives and HR-Pro misses the second action instance.

\begin{figure}[H]
\centering
\begin{subfigure}[b]{0.45\linewidth}
    \includegraphics[width=\linewidth]{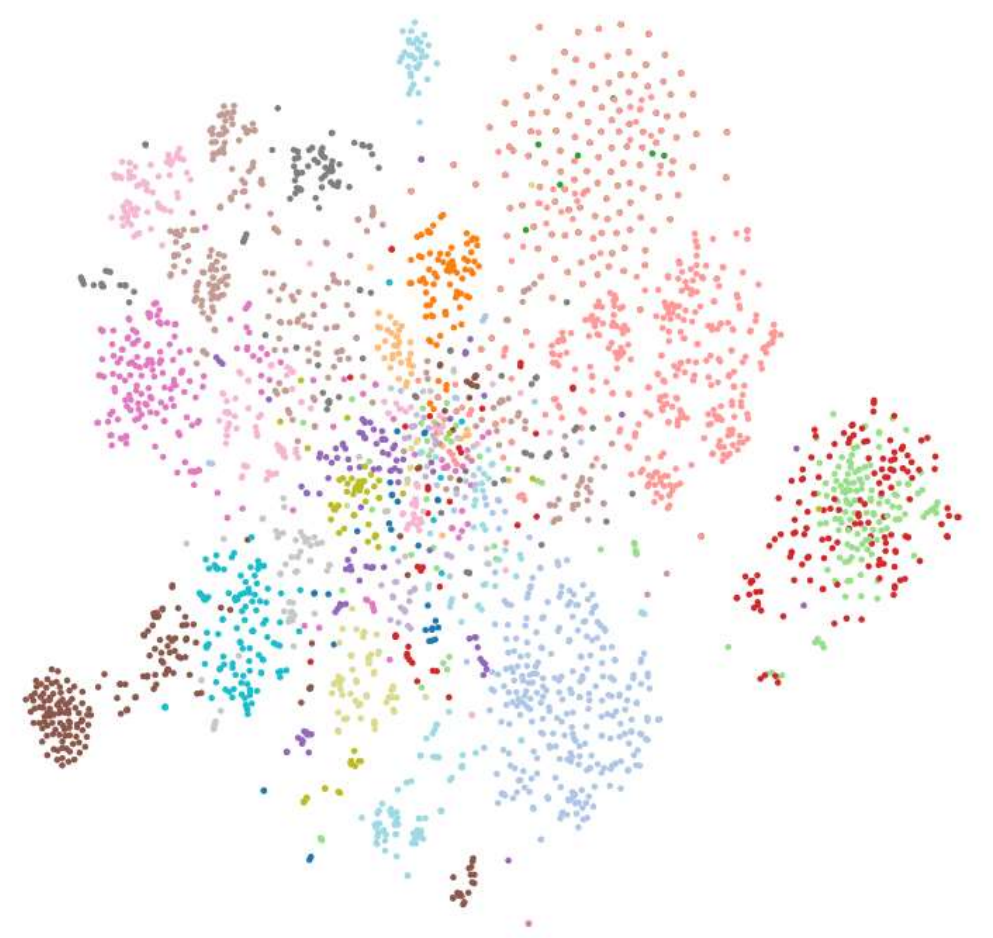}
    \caption{Raw features}
    \label{fig:t-sne_a}
\end{subfigure}
\hfill
\begin{subfigure}[b]{0.45\linewidth}
    \includegraphics[width=\linewidth]{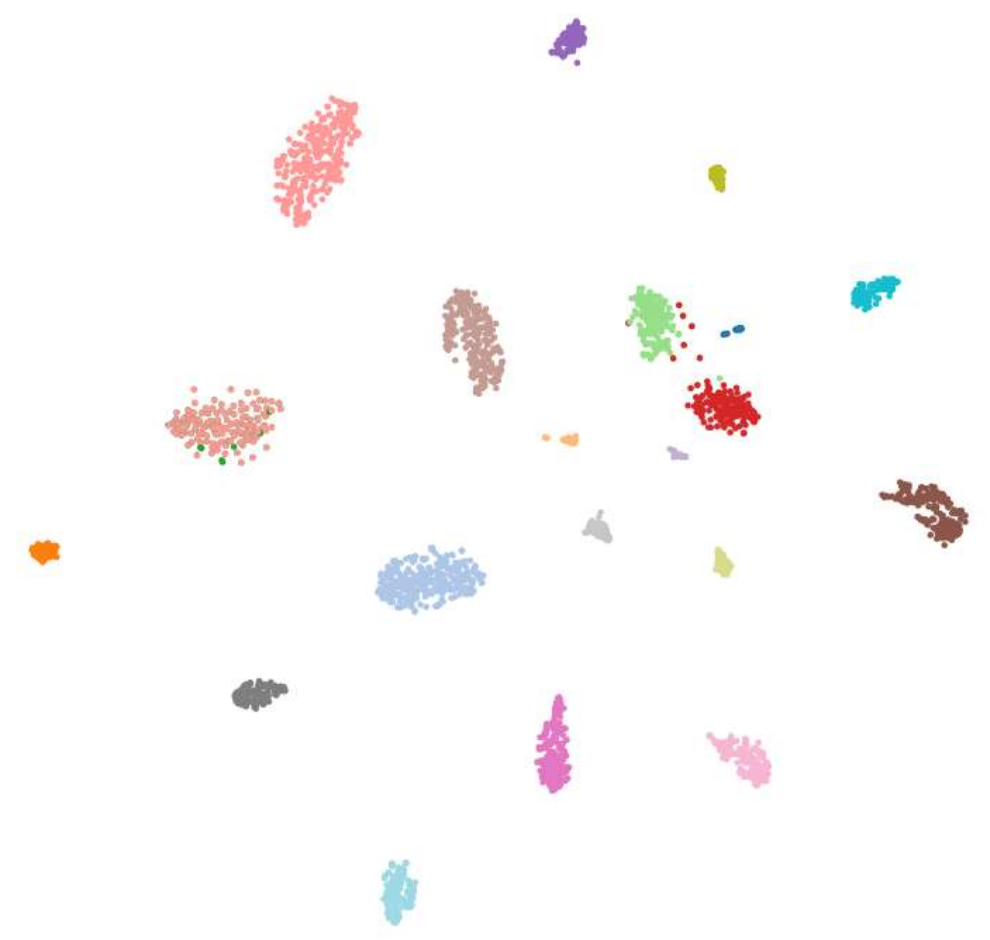}
    \caption{Embedded features}
    \label{fig:t-sne_b}
\end{subfigure}
\caption{t-SNE visualization of the distribution of (a) raw features and (b) embedded features, respectively, where dots represent snippet features and different colors indicate different
action categories.
}
\label{fig:t-sne}
\end{figure}

\noindent\textbf{Feature embedding visualization.}
We use t-SNE to visualize the raw features and embedded features on THUMOS’14 in Figure~\ref{fig:t-sne}.
Figure~\ref{fig:t-sne} (a) shows that the raw features are difficult to distinguish between different actions, but in Figure~\ref{fig:t-sne} (b), they form clearer clusters after training. 

\subsection{Limitations}

\textcolor{black}{
Although our method explicitly models temporal consistency with three self-supervised auxiliary tasks, it still has limitations.}

\begin{figure}[H]
\centering
\includegraphics[width=\linewidth]{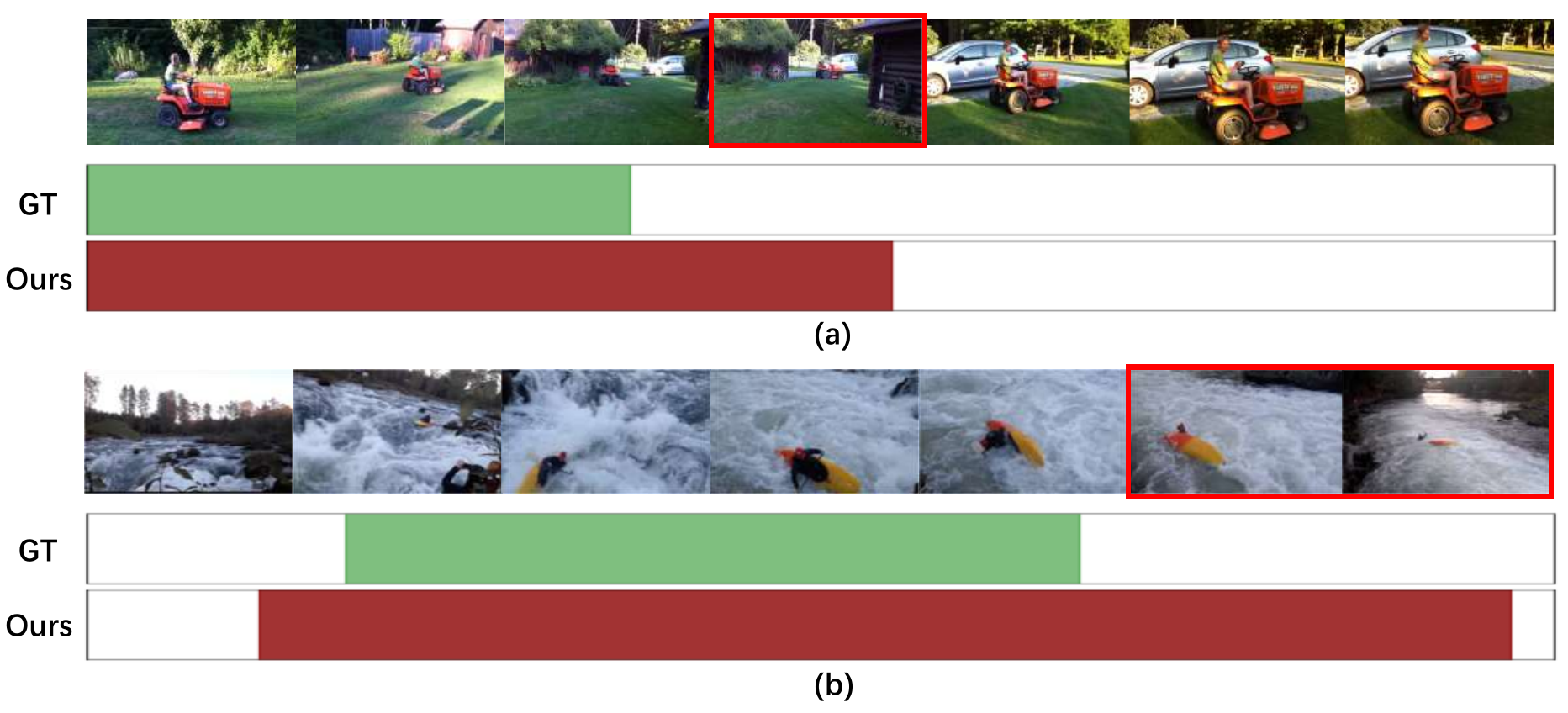}

\caption{Visualization of failure cases. Two representative examples are shown: (a) Cutting the grass and (b) Kayaking. The red boxes indicate false positive predictions, where background segments are incorrectly detected as action instances.
}
\label{fig:failure_cases}

\end{figure}

\textcolor{black}{
Specifically, it may fail to distinguish background segments near action boundaries from true action instances when they share similar visual appearance or contextual information. 
In Figure~\ref{fig:failure_cases}(a), the model correctly localizes the main “Cutting the grass” process, but still predicts the segment after the lawn mower leaves the grass area as “Cutting the grass”. 
In Figure~\ref{fig:failure_cases}(b), the model incorrectly predicts the segment after the kayak overturns as “Kayaking”, although it no longer belongs to a valid action instance. 
These failure cases show that the model still struggles with confusing background segments that are visually or contextually similar to the action.}

\textcolor{black}{
Beyond these failure cases, several limitations deserve further investigation. First, temporally symmetric or cyclic actions may introduce ambiguity into the AOU task, since reversed sequences can be visually or semantically similar to the original ones. Second, the use of a fixed temporal window may limit context coverage for long actions or introduce background noise for short actions. Third, high frame rates may increase the similarity between adjacent frames or snippets, potentially causing temporal leakage in the Action Completion task. In future work, we will explore more adaptive designs, such as adaptive negative sampling, variable-length temporal windows, and adaptive temporal sampling, to further improve robustness.
}

\section{Conclusion}
In this paper, we point out that existing methods mainly rely on point-supervised snippet classification for model training, but lack effective modeling of temporal consistency.
To bridge this gap, we introduce three self-supervised tasks based on point annotations, including {Action Completion}, {Action Order Understanding}, and {Action Regularity Understanding}.
These tasks empower the model to reason about action context, temporal order, and regularity patterns, respectively.
We incorporate these tasks into the base action detection framework to \textcolor{black}{jointly model temporal consistency and improve temporal action localization.}
We will also make the code publicly available.
Extensive experiments on four benchmark datasets demonstrate the robustness and effectiveness of our method and highlight the significant potential of point annotations for temporal action localization.



\bibliographystyle{elsarticle-num}
\bibliography{main}

\end{document}